\documentstyle[psfig]{article}

\begin{document}

\title{Collective Intelligence for Control of  Distributed \\ Dynamical Systems}
\author{D. H. Wolpert, K. R. Wheeler and K. Tumer \\
NASA Ames Research Center, Moffett Field, CA 94035 \\
\\
Tech Report: NASA-ARC-IC-99-44}

\maketitle

\begin{abstract} We consider the El Farol bar problem, also known as
the minority game (W. B. Arthur, {\it The American Economic Review},
84(2): 406--411 (1994), D. Challet and Y.C. Zhang, {\it Physica A},
256:514 (1998)). We view it as an instance of the general problem of
how to configure the nodal elements of a distributed dynamical system
so that they do not ``work at cross purposes'', in that their
collective dynamics avoids frustration and thereby achieves a provided
global goal.  We summarize a mathematical theory for such
configuration applicable when (as in the bar problem) the global goal
can be expressed as minimizing a global energy function and the nodes
can be expressed as minimizers of local free energy functions.  We
show that a system designed with that theory performs nearly optimally
for the bar problem.


\end{abstract}

\pagenumbering{arabic}

\section{Introduction} \label{sec:intro}In many distributed dynamical
systems there is little centralized communication and control among
the individual nodal elements.  Despite this handicap, typically we
wish to design the system so that its dynamical behavior has some desired
form. Often the quality of that behavior can be expressed as a
(potentially path-dependent) global energy function, $G$. The
associated design problem is particularly interesting when we can also
express the individual nodal elements $\eta$ as minimizers of
``local'' energy functions $\gamma_\eta$. Given $G$, this reduces the
problem to determining the optimal associated
\{$\gamma_\eta$\}. 

Because the argument lists of the $\gamma_\eta$ may overlap, what
action $\eta$ should take at time $t$ to minimize $\gamma_\eta$ may
depend on what actions the other nodes take at $t$. Since without
binding contracts $\eta$ cannot know those other actions ahead of
time, it cannot assuredly minimize $\gamma_\eta$ in general. We are
particularly interested in cases where each $\eta$ addresses this
problem by using machine-learning (ML) techniques to determine its
actions. (In its use of such techniques that trade off exploration and
exploitation, such an $\eta$ often approximates a stochastic node
following the distribution that minimizes $\eta$'s free energy --- see
below.) In such cases the challenge is to choose the \{$\gamma_\eta$\}
so that the associated system of good (but suboptimal) ML-based nodes
induces behavior that best minimizes the provided function $G$.


We refer to a system designed this way, or more generally to a system
investigated from this perspective, as a COllective INtelligence
(COIN)~\cite{wotu99b,wotu99a}. To agree with bar problem and
game-theory terminology, we refer to the nodes as $agents$, $G$ as
(minus) {\it world utility}, and the \{$\gamma_\eta$\} as (minus) {\it
private utilities}. As an example of this terminology, a spin glass in
which each spin $\eta$ is at an energy minimum given the states of the
other spins is a ``Nash equilibrium'' of an associated ``game''
~\cite{futi91}, a game formed by identifying each agent with a spin
$\eta$ and its associated private utility function with $\eta$'s
energy function.

Arthur's bar problem~\cite{arth94} can be viewed as a problem in
designing COINs.  Loosely speaking, in this problem at each time $t$
each agent $\eta$ decides whether to attend a bar by predicting, based
on its previous experience, whether the bar will be too crowded to be
``rewarding'' at that time, as quantified by a reward function
$R_{UD;\eta}$.  The greedy nature of the agents frustrates the
global goal of maximizing $G = \sum_\eta R_{UD;\eta}$ at $t$. This is
because if most agents think the attendance will be low (and therefore
choose to attend), the attendance will actually be high, and
vice-versa.  This frustration effect makes the bar problem
particularly relevant to the study of the physics of emergent behavior
in distributed
systems~\cite{caga99,chzh98,chen97,capl98,joja98,sama97,zhan98}.


In COIN design we try to avoid such effects by determining new
utilities \{$\gamma_\eta$\} so that all agents trying to minimize
those new utilities means that $G$ is also minimized. (Of course, we
wish to determine the \{$\gamma_\eta$\} without first explicitly
solving for the minimum of $G$.)  As an analogy, economic systems
sometimes have a ``tragedy of the commons'' (TOC) ~\cite{hard68},
where each
agent's trying maximize its utility results in collective behavior
that $minimizes$ each agent's utility, and therefore minimizes
$minimizes$ $G = \sum_\eta R_{UD;\eta}$. One way the TOC is avoided in
real-world economies is by reconfiguring the agents' utility functions
from \{$R_{UD;\eta}$\} to a set of \{$\gamma_\eta$\} that results in
better $G$, for example via punitive legislation like anti-trust
regulations. Such utility modification is exactly the approach used in
COIN design.


We recently applied such COIN design to network packet
routing~\cite{wotu99a}.  In conventional packet routing each router
uses a myopic shortest path algorithm (SPA), with no concern for
side-effects of its decisions on an external world utility like global
throughput (e.g., for whether those decisions induce bottlenecks).  We
found that a COIN-based system has significantly better throughput
than does a conventional SPA~\cite{wotu99a}, even when the agents in
that system had to predict quantities (e.g., delays on links) that
were directly provided to the SPA.

In this paper we confront frustration effects more directly, in the
context of the bar problem. In the next section we present (a small
portion of) the theory of COINs. Then we present experiments applying
that theory to the distributed control of the agents in the bar
problem. Those experiments indicate that by using COIN theory we can
avoid the frustration in the bar problem and thereby achieve almost
perfect minimization of the global energy.

\section{Theory of COINs} \label{sec:math} We consider the state of
the system across a set of consecutive time steps, $t \in \{0, 1,
...\}$.  Without loss of generality, all relevant characteristics of
agent $\eta$ at time $t$ --- including its internal parameters at that
time as well as its externally visible actions --- are encapsulated by
a Euclidean vector $\underline{\zeta}_{\eta,t}$, the $state$ of agent
$\eta$ at time $t$. $\underline{\zeta}_{,t}$ is the set of the states
of all agents at $t$, and $\underline{\zeta}$ is the state of all
agents across all time.

So {\bf world utility} is $G(\underline{\zeta})$, and when $\eta$ is
an ML algorithm ``striving to increase'' its {\bf private utility}, we
write that utility as $\gamma_{\eta}(\underline{\zeta})$.  The
mathematics is generalized beyond such ML-based agents through an
artificial construct: the {\bf personal utilities}
\{$g_{\eta}(\underline{\zeta})$\}. We restrict attention to utilities
of the form $\sum_t R_t(\underline{\zeta}_{,t})$ for {\bf reward
functions} $R_t$.



We are interested in systems whose dynamics is deterministic.  (This
covers in particular any system run on a digital computer.)  We
indicate that dynamics by writing $\underline{\zeta} =
C(\underline{\zeta}_{,0})$. So all characteristics of an agent $\eta$
at $ t = 0$ that affects the ensuing dynamics of the system, including
in particular its private utility if it has one, must be included in
$\underline{\zeta}_{\eta,0}$.

{\bf Definition:} A system is {\bf factored} if for each agent
$\eta$ individually,
\begin{equation}
g_\eta(C(\underline{\zeta}_{,0})) \geq 
  g_\eta(C(\underline{\zeta}'_{,0})) 
\; \; \; \Leftrightarrow \; \; \;
G(C(\underline{\zeta}_{,0})) \geq 
  G(C(\underline{\zeta}'_{,0})) \; ,
\end{equation}
for all pairs $\underline{\zeta}_{,0}$ and $\underline{\zeta}'_{,0}$ that differ 
only for node $\eta$.


For a factored system, the side effects of a change to $\eta$'s $t =
0$ state that increases its personal utility cannot decrease world
utility. If the separate agents have high personal utilities, by luck
or by design, then they have not frustrated each other, as far as
$G$ is concerned.

The definition of factored is carefully crafted. In particular, it
does {\it not} concern changes in the value of the utility of agents
other than the one whose state is varied. Nor does it concern changes
to the states of more than one agent at once.  Indeed, consider the
following alternative desideratum to having the system be factored:
any change to $\underline{\zeta}_{,0}$ that simultaneously improves all
agents' ensuing utilities must also improve world utility.  Although
it seems quite reasonable, there are systems that obey this
desideratum and yet quickly evolve to a {\it minimum} of world
utility.  For example, any system that has $G(\underline{\zeta}) =
\sum_{\eta}g_{\eta}(\underline{\zeta})$ obeys this desideratum, and
yet as shown below, such systems entail a TOC in the the bar
problem.

For a factored system, when every agents' personal utility is
optimizal, given the other agents' behavior, world utility is at a
critical point~\cite{wotu99b}.  In game-theoretic terms, optimal
global behavior corresponds to the agents' reaching a personal utility
Nash equilibrium for such systems~\cite{futi91}.  Accordingly, there
can be no TOC for a factored system.

As a trivial example, if $g_{\eta} = G \, \, \forall \eta$, then the
system is factored, regardless of $C$. However there exist other,
often preferable sets of \{$g_\eta$\}, as illustrated in the following
development.

{\bf Definition:} The ($t=0$) {\bf effect set} of node $\eta$ at 
$\underline{\zeta}$, $C^{eff}_\eta(\underline{\zeta})$, is the set of 
all components $\underline{\zeta}_{\eta',t'}$ for which
$\partial_{\underline{\zeta}_{\eta,0}}
(C(\underline{\zeta}_{,0}))_{\eta',t'} \neq \vec{0}$.
$C^{eff}_\eta$ with no specification of 
$\underline{\zeta}$ is defined as $\cup_{\underline{\zeta} \in C}
C^{eff}_\eta(\underline{\zeta})$.


{\bf Definition:} Let $\sigma$ be a set of agent-time pairs.
$\mbox{CL}_{\eta}(\underline{\zeta})$ 
is $\underline{\zeta}$ modified by ``clamping'' the states
corresponding to all elements of $\sigma$ to some arbitrary pre-fixed value, here 
taken to be $\vec{0}$. The {\bf wonderful life utility} (WLU) for 
$\sigma$ at $\underline{\zeta}$ is defined as:
\begin{equation}
WLU_{\sigma}(\underline{\zeta}) \equiv G(\underline{\zeta}) -
G(\mbox{CL}_{\sigma} (\underline{\zeta})) \; .
\end{equation}
\noindent
In particular, the  WLU for the effect set of node $\eta$ is
$G(\underline{\zeta}) - G(\mbox{CL}_{C^{eff}_\eta} (\underline{\zeta}))$.

$\eta$'s effect set WLU is analogous to the change world utility would
undergo had node $\eta$ ``never existed''.  (Hence the name of this
utility - cf. the Frank Capra movie.)  However $\mbox{CL}(.)$ is a
purely ``fictional'', counter-factual mapping, in that it produces a
new $\underline{\zeta}$ without taking into account the system's
dynamics.  The sequence of states produced by the clamping operation
in the definition of the WLU need not be consistent with the dynamical
laws embodied in $C$.  This is a crucial strength of effect set WLU.
It means that to evaluate that WLU we do {\it not} try to infer how
the system would have evolved if node $\eta$'s state were set to
$\vec{0}$ at time 0 and the system re-evolved. So long as we know $G$
and the full $\underline{\zeta}$, and can accurately estimate what
agent-time pairs comprise $C^{eff}_\eta$, we know the value of
$\eta$'s effect set WLU --- even if we know nothing of the details of
the dynamics of the system.

{\bf Theorem 1:} A COIN is factored if $g_\eta = WLU_{C^{eff}_\eta} \;
\forall \eta$  (proof in \cite{wotu99b}).

If our system is factored with respect to personal utilities
\{$g_\eta$\}, then we want each $\underline{\zeta}_{\eta,0}$ to be a
state with as high a value of $g_\eta(C(\underline{\zeta}_{,0}))$ as
possible. Assuming $\eta$ is ML-based and able to achieve close to the
largest possible value of any private utility specified in
$\underline{\zeta}_{\eta,0}$, we would likely be in such a state of
high personal utility if $\eta$'s private utility were set to the
associated personal utility: $\gamma_\eta \equiv
\underline{\zeta}_{\eta,0;{\rm private-utility}} = g_\eta$. Enforcing
this equality, our problem becomes determining what \{$\gamma_\eta$\}
the agents will best be able to maximize while also causing dynamics
that is factored with respect to the \{$\gamma_\eta$\}.

Now regardless of $C(.)$, both $\gamma_\eta = G \; \forall \eta$ and
$\gamma_\eta = WLU_{C^{eff}_\eta} \; \forall \eta$ are factored
systems (for $g_\eta = \gamma_\eta$). However since each agent is
operating in a large system, it may experience difficulty discerning
the effects of its actions on $G$ when $G$ sensitively depends on all
components of the system. Therefore each $\eta$ may have difficulty
learning how to achieve high $\gamma_\eta$ when $\gamma_\eta = G$.
This problem can be obviated using effect set WLU as the private
utility, since the subtraction of the clamped term removes some of the
``noise'' of the activity of other agents, leaving only the underlying
``signal'' of how the agent in question affects the utility.

We can quantify this signal/noise effect by comparing the ramifications on the
private utilities arising from changes to $\underline{\zeta}_{\eta,0}$
with the ramifications arising from changes to
$\underline{\zeta}_{\;\hat{}\eta,0}$, where $\;\hat{}\eta$ represents
all nodes {\em other} than $\eta$.  We call this quantification the
{\bf learnability} $\lambda_{\eta,\gamma_\eta}(\underline{\zeta})$:
\begin{equation}
\lambda_{\eta,\gamma_\eta}(\underline{\zeta}) \equiv \frac{ \|
\vec{\nabla}_{\underline{\zeta}_{\eta,0}}
\gamma_\eta(C(\underline{\zeta}_{,0}))\| } {\|
\vec{\nabla}_{\underline{\zeta}_{\; \hat{}\eta,0}}
\gamma_\eta(C(\underline{\zeta}_{,0}))\| } \;.  \label{eq:learn}
\end{equation} \noindent

\vspace*{-.1in}

\noindent

{\bf Theorem 2:} Let $\sigma$ be a set
containing $C^{eff}_{\eta}$. Then
\begin{eqnarray*}
\frac{\lambda_{\eta,WLU_{\sigma}}(\underline{\zeta})}
     {\lambda_{\eta,G}(\underline{\zeta})} =
\frac{\| \vec{\nabla}_{\underline{\zeta}_{\;\hat{}\eta,0} }
	G(C(\underline{\zeta}_{,0})) }
     {\| \vec{\nabla}_{\underline{\zeta}_{\;\hat{}\eta,0} }
	G(C(\underline{\zeta}_{,0})) -
        \vec{\nabla}_{\underline{\zeta}_{\;\hat{}\eta,0} }
	G(\mbox{CL}_{\sigma}(C(\underline{\zeta}_{,0}))) \|} {\rm
proof} \; {\rm in} \;  \cite{wotu99b}.
\end{eqnarray*}
This ratio of gradients should be large whenever $\sigma$ is a small
part of the system, so that the clamping won't affect $G$'s dependence
on $\underline{\zeta}_{\;\hat{}\eta,0}$ much, and therefore that
dependence will approximately cancel in the denominator term. In such
cases, WLU will be factored just as $G$ is, but far more
learnable. The experiments presented below illustrate the power of
this fact in the context of the bar problem, where one can readily
approximate effect set WLU and therefore use a utility for which the
conditions in Thm.'s 1 and 2 should approximately hold.


\section{Experiments} \label{sec:bar} We modified Arthur's original
problem to be more general, and since we are not interested here in
directly comparing our results to those
in~\cite{arth94,cama97,chzh98,capl98}, we use a more conventional ML
algorithm than the ones investigated
in~\cite{arth94,cama97,chzh98,capl98,sebe98}, an algorithm that
approximately minimizes free energy. These modifications are similar
to those in~\cite{caga99}.

There are $N$ agents, each picking one of seven nights to attend
a bar the following week, a process that is then repeated.  In each
week, each agent's pick is determined by its predictions of the
associated rewards it would receive.  Each such prediction in turn is
based solely upon the rewards received by the agent in those preceding
weeks in which it made that pick.

The world utility is $G(\underline{\zeta})= \sum_t
R_G(\underline\zeta_{,t})$, where $R_G(\underline\zeta_{,t}) \equiv
\sum_{k=1}^7 \phi_k(x_k(\underline{\zeta},t))$,
$x_k(\underline{\zeta}, t)$ is the total attendance on night $k$ at
week $t$, $\phi_k(y) \equiv \alpha_k y \exp{(-y/c)}$; and $c$ and the
\{$\alpha_k$\} are real-valued parameters.  Intuitively, this $G$ is
the sum of the ``world rewards'' for each night in each week.  Our
choice of $\phi_k(.)$ means that when too few agents attend some night
in some week, the bar suffers from lack of activity and therefore the
world reward is low.  Conversely, when there are too many agents the
bar is overcrowded and the reward is again low.


Two different $\vec{\alpha}$'s are investigated.  One treats all
nights equally; $\vec{\alpha}~=~[1~1~1~1~1~1~1]$.  The other is only
concerned with one night; $\vec{\alpha}~=~[0~0~0~7~0~0~0]$.  $c = 6$
and $N$ is 4 times the number of agents needed to allow $c$ agents to
attend the bar on each of the seven nights, i.e., there are $4 \times
6 \times 7 = 168$ agents. For the purposes of the CL operation, an
agent's action at time $t$ is represented as a unary seven-dimensional
vector, so the ``clamped pick'' is (0,0,0,0,0,0,0).

Each $\eta$ has a 7-dimensional vector representing its estimate of
the reward it would receive for attending each night of the week.  At
the end of each week, the component of this vector corresponding to
the night just attended is proportionally adjusted towards the actual
reward just received.  At the beginning of the succeeding week, to
trade off exploration and exploitation, $\eta$ picks the night to
attend randomly using a Boltzmann distribution with 7 energies
$\epsilon_i(\eta)$ given by the components of $\eta$'s estimated
rewards vector, and with a temperature decaying in time.  This
distribution of course minimizes the expected free energy of $\eta$,
$E(\epsilon(\eta)) - TS$, or equivalanetly maximizes entropy $S$
subject to having expected energy given by $T$. This learning
algorithm is similar to Claus and Boutilier's independent learner
algorithm ~\cite{clbo98}.

We considered three agent reward functions, using the same learning
parameters (learning rate, Boltzmann temperature, decay rates, etc.)
for each.  The first reward function had $\gamma_{\eta} = G \, \,
\forall \eta$, i.e., agent $\eta$'s reward function equals $R_G$.
The other two reward functions are:
\begin{eqnarray*}
   \mbox{\rm Uniform Division (UD):} & 
   R_{UD;\eta}(\underline{\zeta}_{,t})  & \equiv
   \phi_{d_\eta} (x_{d_\eta} (\underline{\zeta},t))/x_{d_\eta}(\underline{\zeta},t) \\ [-.1in]
   \mbox{Wonderful Life (WL):} \; \; \;  & 
   R_{WL;\eta}(\underline{\zeta}_{,t}) & \equiv 
	R_G(\underline{\zeta}_{,t}) -
	R_G(CL_\eta(\underline{\zeta}_{,t})) \; ,
\end{eqnarray*}
\renewcommand{\arraystretch}{1.0}
\noindent
where $d_{\eta}$ is the night picked by $\eta$.
The original version of the bar problem in the physics literature~\cite{chzh98} is
a special case where there are two ``nights'' in the week (one of
which corresponds to ``staying at home''); $\vec{\alpha}$ is
uniform; $\phi_k(x_k) = min_i(x_i) \delta_{k,argmin_i(x_i)}$; and $R_{UD\eta}$
is used.

The conventional $R_{UD}$ reward is a ``natural'' reward function to
use; each night's total reward is uniformly divided among the agents
attending that night.  In particular, if $g_{\eta} = \gamma_\eta
\equiv \sum_t R_{UD;\eta}(\underline{\zeta},t)$, $G(\underline{\zeta})
= \sum_{\eta} g_{\eta}(\underline{\zeta})$, so the ``alternative
desideratum'' discussed above is met. In contrast, $R_G$ results in
the system meeting the desideratum of factoredness. $R_G$ suffers from
poor learnability, at least in comparison to that of $R_{WL}$; by
Eq.~\ref{eq:learn} the ratio of learnabilities is approximately 11
(see \cite{wotu99b} for details).  As another point of comparison, to
evaluate $R_{WL}$ each agent only needs to know the total attendance
on the night it attended, unlike with $R_G$, which requires
centralized communication concerning all 7 nights.

Finally, in the bar problem the only interaction between any pair of
agents is indirect, via small effects on each others' rewards; each
$\eta$'s action at time $t$ has its primary effect on $\eta$'s own
future actions. So the effect set of $\eta$'s entire sequence of
actions is well-approximated by $\underline{\zeta}_{\eta,}$. In turn,
since that sequence is all that is directly affected by the choice of
$\eta$'s private utility, the effect set of
$\underline{\zeta}_{\eta,0;{\rm private-utility}}$ can be approximated
by $\underline{\zeta}_{\eta,}$, and therefore so can the effect set of
the full $\underline{\zeta}_{\eta,0}$. Therefore we can approximate
the effect set WLU for $\eta$ as $\sum_t R_{WLU;\eta}
(\underline{\zeta}_{\eta,t;{\rm pick}})$. So we expect that use of
$R_{WLU;\eta}$ should result in (close to) factored dynamics.

\begin{figure}[ht]
	\vspace*{-.1in}
   \centerline{\mbox{
   	\psfig{figure={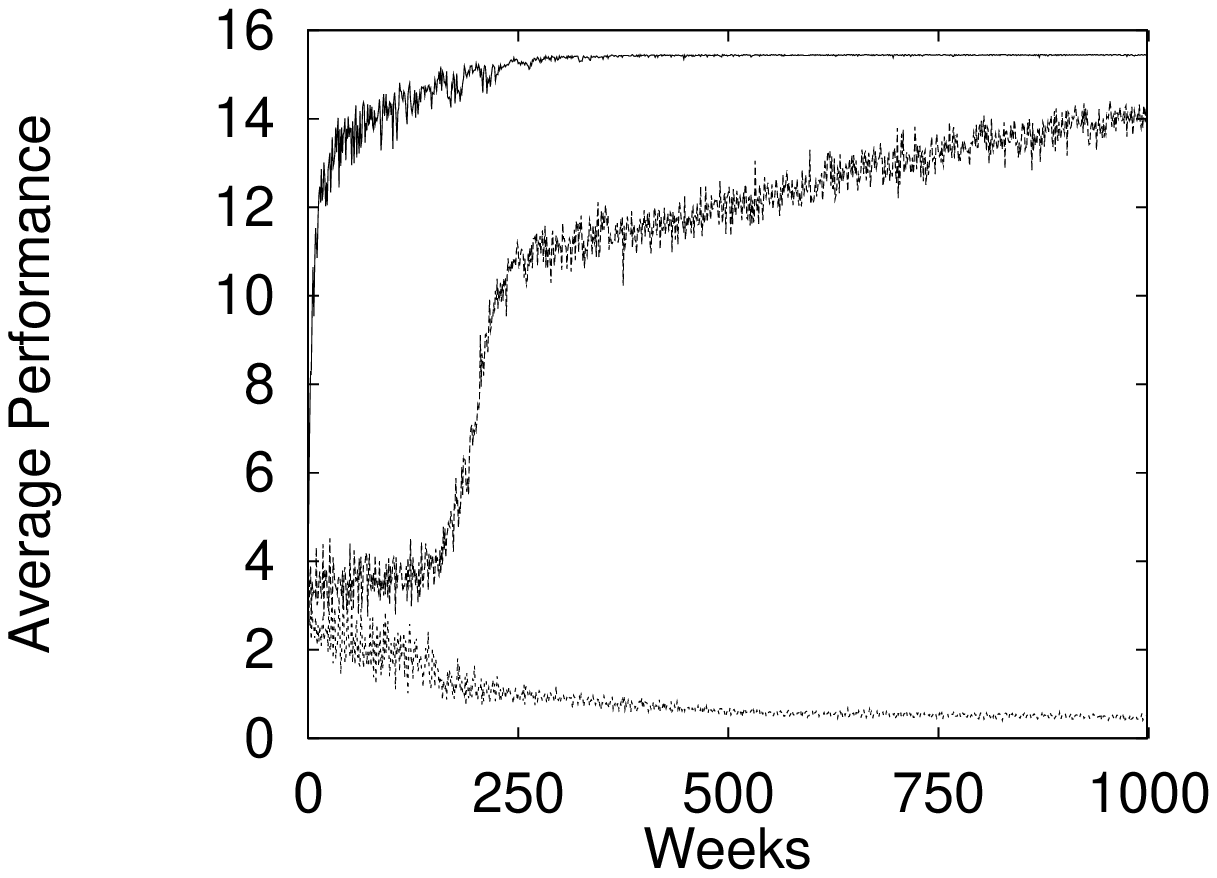},width=2.70in,height=1.8in}
	\hspace{0.10in}
   	\psfig{figure={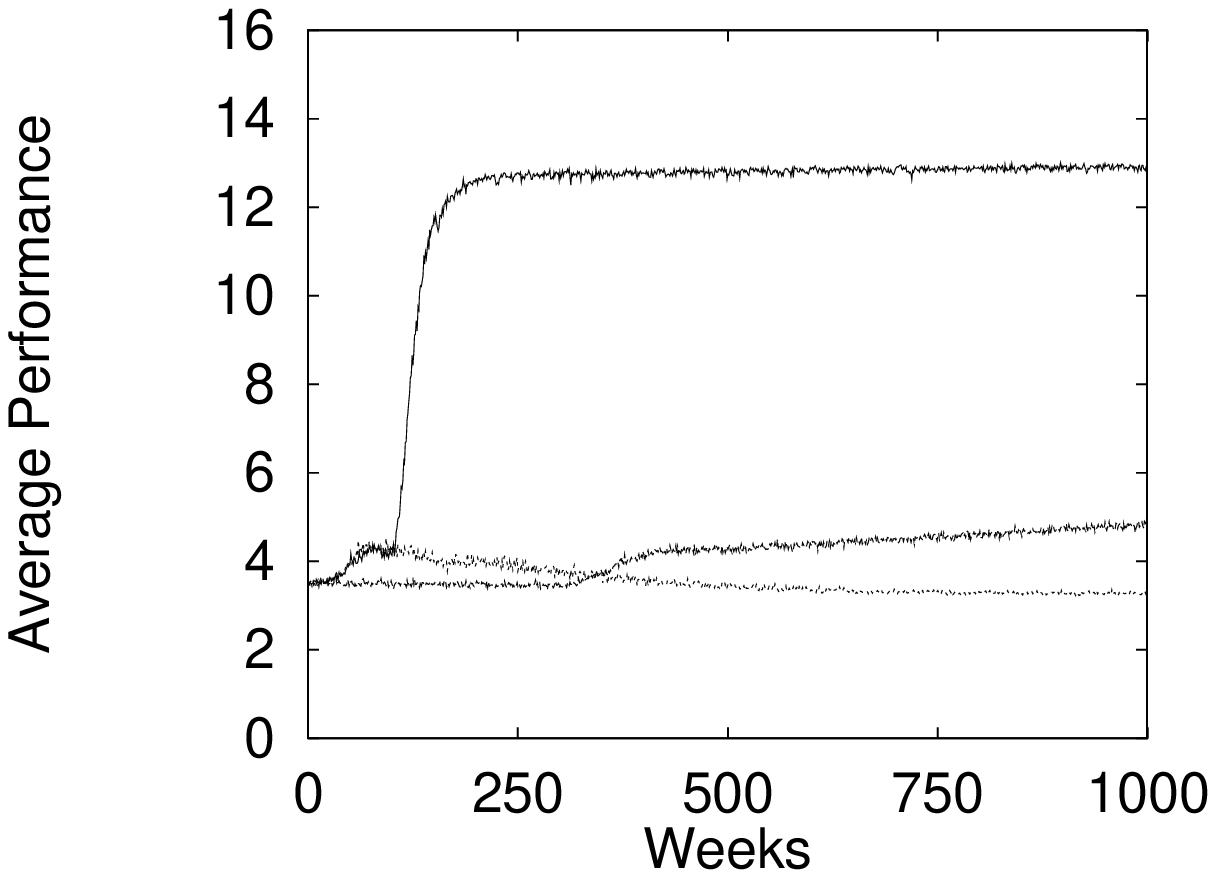},width=2.70in,height=1.8in}
   }}
	\vspace*{-.1in}
  \caption{Average world reward when $\vec{\alpha}~=~[0~0~0~7~0~0~0]$ (left) and
	when $\vec{\alpha}~=~[1~1~1~1~1~1~1]$ (right). In both plots the top 
	curve is $R_{WL}$, middle is $R_G$, and bottom is $R_{UD}$.}
\label{fig:barfig}
\end{figure}

Figure~\ref{fig:barfig} graphs world reward value as a function of
time, averaged over 50 runs, for all three reward functions, for both
$\vec{\alpha}~=~[1~1~1~1~1~1~1]$ and $\vec{\alpha}~=~[0~0~0~7~0~0~0]$.
Performance with $R_G$ eventually converges to the global optimum.
This agrees with the results obtained by Crites~\cite{crba96} for the
bank of elevators control problem.  Systems using $R_{WL}$ also
converged to optimal performance.  This indicates that in the bar
problem $\gamma_\eta$'s effect set is sufficiently well-approximated
by $\eta$'s future actions so that the conclusions of theorems 1 and 2
hold.

However since the $R_{WL}$ reward has better ``signal to noise'' than
than the $R_G$ reward (see above), 
convergence with $R_{WL}$ is far quicker than with $R_G$. 
Indeed, when $\vec{\alpha}~=~[0~0~0~7~0~0~0]$,
systems using $R_G$ converge in 1250 weeks, which is
5 times worse than the systems using $R_{WL}$.  When
$\vec{\alpha}~=~[1~1~1~1~1~1~1]$  systems take 6500
weeks to converge with $R_G$, which is more than {\it 30 times}
worse than the time with $R_{WL}$.  This slow convergence of
systems using $R_G$ is a result of the reward signal being
``diluted'' by the large number of agents in the system.

\begin{figure}[ht]
	\vspace*{-.1in}
   \centerline{\mbox{
   	\psfig{figure={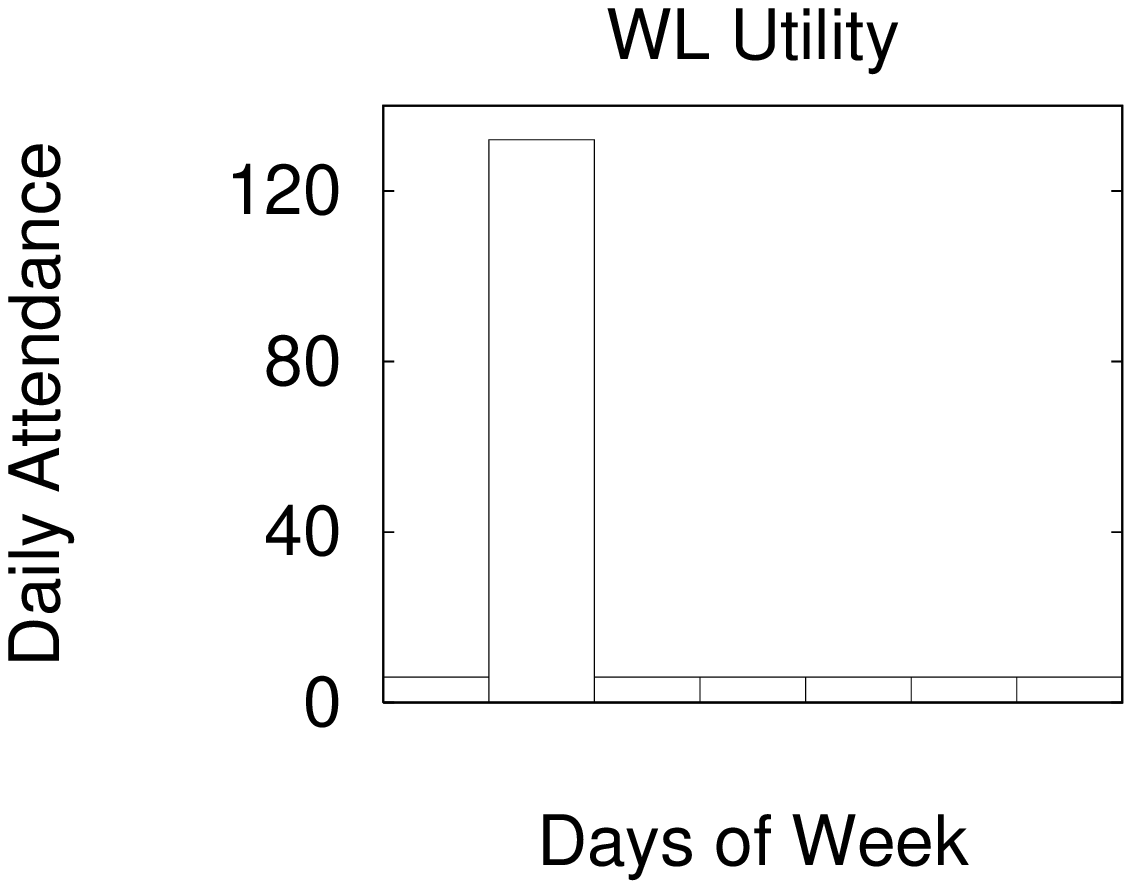},width=1.9in,height=1.5in}
	\psfig{figure={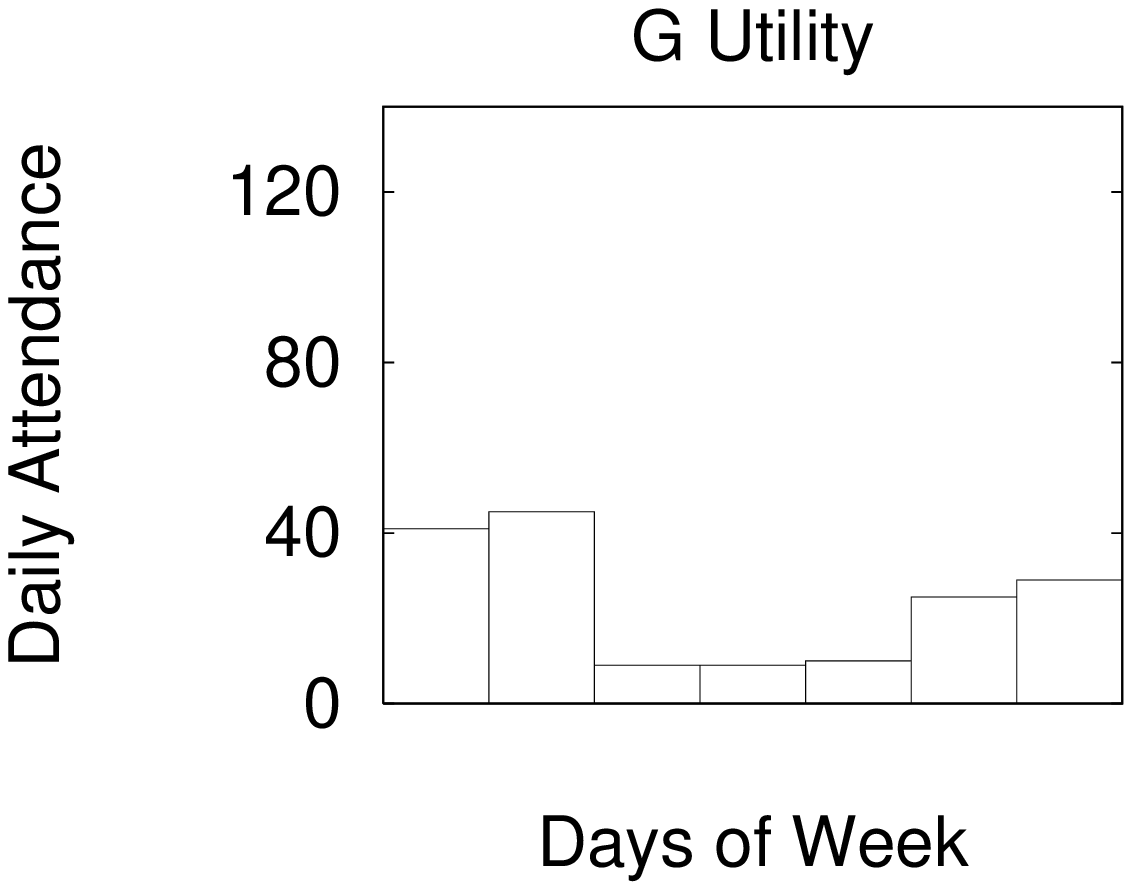},width=1.9in,height=1.5in}
\psfig{figure={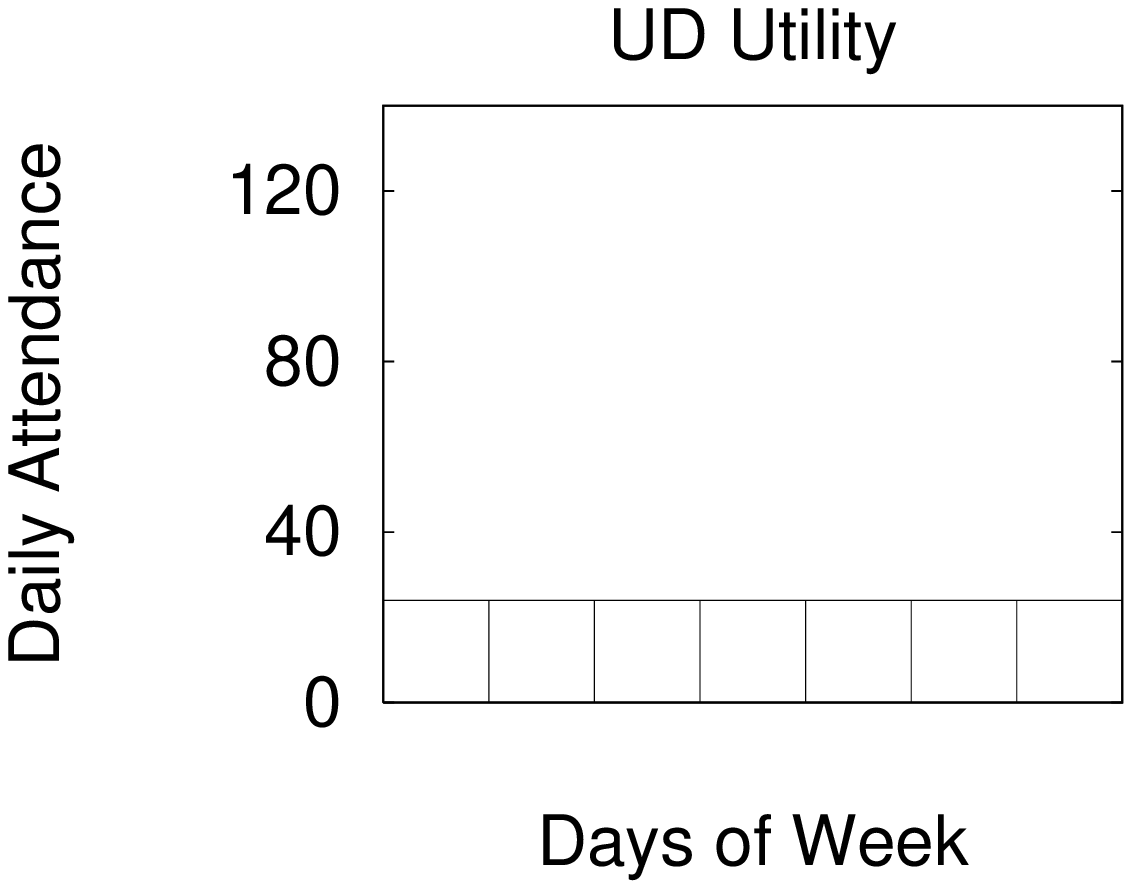},width=1.9in,height=1.5in}
   }}
	\vspace*{-.1in}
   \caption{Typical daily attendance when $\vec{\alpha}~=~[1~1~1~1~1~1~1]$ for 
   $R_{WL}$, $R_G$, and $R_{UD}$, respectively.} 
\label{fig:attend}
\end{figure}

In contrast to the behavior for COIN theory-based reward functions,
use of conventional $R_{UD}$ reward results in very poor world reward
values that deteriorated with time. This is an instance of the
TOC. For example, when $\vec{\alpha}~=~[0~0~0~7~0~0~0]$, it is in
every agent's interest to attend the same night --- but their doing so
shrinks the world reward ``pie'' that must be divided among all
agents.  A similar TOC occurs when $\vec{\alpha}$ is uniform.  This is
illustrated in fig.~\ref{fig:attend} which shows a typical example of
\{$x_k(\underline{\zeta}, t)$\} for each of the three reward functions
for $t~=~2000$.  In this example using $R_{WL}$ results in optimal
performance, with 6 agents each on 6 separate nights, and the
remaining 132 agents on one night (average world reward of 13.05).  In
contrast, $R_{UD}$ results in a uniform distribution of agents and has
the lowest average world reward (3.25). Use pf $R_G$ results in an
intermediate average world reward (6.01).

\vspace*{-.1in}

\begin{figure}[ht]
	\vspace*{-.1in}
   \centerline{\mbox{
   	\psfig{figure={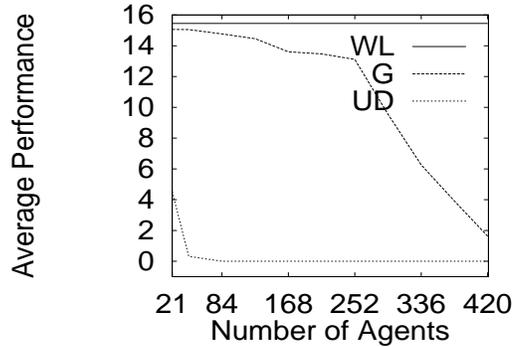},width=2.8in,height=1.8in}
   }}
	\vspace*{-.1in}
   \caption{Behavior of each reward function with respect to the number of
   agents for  $\vec{\alpha}~=~[0~0~0~7~0~0~0]$.} 
\label{fig:numagents}
\end{figure}

\vspace*{-.1in}

Figure~\ref{fig:numagents} shows how performance at $t = 2000$ scales
with $N$ for each reward function for
$\vec{\alpha}~=~[0~0~0~7~0~0~0]$.  Systems using $R_{UD}$ perform
poorly regardless of $N$. Systems using $R_G$ perform well when $N$ is
low. As $N$ increases however, it becomes increasingly difficult for
the agents to extract the information they need from $R_G$. (This
problem is significantly worse for uniform $\vec{\alpha}$.)  Systems
using $R_{WL}$ overcome this learnability problem because $R_{WL }$ is
based on clamping of all agents but one, and therefore is not
appreciably affected by $N$.

\section{Conclusion} \label{sec:conc} The theory of COINs is concerned
with distributed systems of controllers in which each controller
strives to minimize an associated local energy function. That theory
suggest how to initialize and then update those local energy functions
so that the resultant global dynamics will achieve a global goal. In
this paper we present a summary of the part of that theory dealing with
how to initialize the local energy functions.  We present experiments
applying that theory to the control of individual agents in difficult
variants of Arthur's El Farol Bar problem.  In those experiments, the
COINs quickly achieve nearly optimal performance, in contrast to the
other systems we investigated.  This
demonstrates that even when the conditions required by the
initialization theorems of COIN theory do not hold exactly, they often
hold well enough so that they can be applied with confidence.
In particular the COINs automatically
avoid the tragedy of the commons inherent in the bar problem.

%

\bibliographystyle{plain}

\end{document}